\documentclass{article}
\usepackage{ijcai20}

\usepackage{times}

\usepackage{soul}
\usepackage{url}
\usepackage[hidelinks]{hyperref}
\usepackage[utf8]{inputenc}
\usepackage[small]{caption}
\usepackage{graphicx}
\usepackage{amsmath}
\usepackage{booktabs}
\usepackage{multirow}
\usepackage{dirtytalk}
\title{Interpretable Crowd Flow Prediction with Spatial-Temporal Self-Attention}

\author{
Haoxing Lin$^{1,2}$\and
Weijia Jia$^{1,2,3}$\footnote{Contact Author}\and
Yongjian You$^{1,3}$\and
Yiping Sun$^{1,3}$\\
\affiliations
$^1$State Key Lab of Internet of Things for Smart City, 
$^2$FST, University of Macau\\
$^3$Shanghai Jiaotong University\\
\emails
\{mb85410, jiawj\}@um.edu.mo,
\{youyongjian, sunacc\}@sjtu.edu.cn,
}

\begin{document}

\maketitle

\begin{abstract}
    Crowd flow prediction has been increasingly investigated in intelligent urban computing field as a fundamental component of urban management system. The most challenging part of predicting crowd flow is to measure the complicated spatial-temporal dependencies. A prevalent solution employed in current methods is to divide and conquer the spatial and temporal information by various architectures (e.g., CNN/GCN, LSTM). However, this strategy has two disadvantages: (1) the sophisticated dependencies are also divided and therefore partially isolated; (2) the spatial-temporal features are transformed into latent representations when passing through different architectures, making it hard to interpret the predicted crowd flow. To address these issues, we propose a Spatial-Temporal Self-Attention Network (\textbf{STSAN}) with an ST encoding gate that calculates the entire spatial-temporal representation with positional and time encodings and therefore avoids dividing the dependencies. Furthermore, we develop a Multi-aspect attention mechanism that applies scaled dot-product attention over spatial-temporal information and measures the attention weights that explicitly indicate the dependencies. Experimental results on traffic and mobile data demonstrate that the proposed method reduces inflow and outflow RMSE by 16\% and 8\% on the Taxi-NYC dataset compared to the SOTA baselines. Codes: https://github.com/starkfather/STSAN
\end{abstract}

\section{Introduction}

Crowd flow prediction has drawn increasing attention in AI research field because of its critical role in urban management system. Since high-level applications such as intelligent resource allocation and dynamic traffic management rely heavily on crowd flow prediction, its effectiveness and interpretability become very crucial. While a substantial amount of crowd flow data has been generated, deep learning approaches have been increasingly investigated and have outperformed the traditional methods.
    
Given historical observations, crowd flow prediction means to predict the volumes of crowd flows in the upcoming timestamp. Since deep learning methods obtained significant advantages in modeling both spatial and temporal dependencies \cite{Deeplearning}, deep residual network \cite{DeepResidual}, graph convolution network \cite{gcn}, and recurrent neural network \cite{ConvLSTM} dominate the crowd flow prediction field. Several works apply deep residual networks to capture spatial dependencies from different periodic sequences \cite{stresnet,ZHANG_TKDE}, while some others handle the spatial or graph convolutional results with LSTM \cite{lstm} to capture the temporal dependencies \cite{stdn,st_mgcn}. However, even dividing the spatial and temporal information and conquering them with particular techniques reduce the complexity and maximize the capability of each measurement, the sophisticated spatial-temporal dependencies are also divided. Other than that, as the spatial-temporal information is transformed into latent representation to go through heterogeneous architectures, the dependencies are measured implicitly, which outputs only the predicted values without telling users where the crowd flows come from and which historical timestamp is most relevant.

Two main reasons urge current methods to employ the divide-and-conquer strategy. First, since the spatial-temporal information has at least three dimensions, the feature space can grow massively if the considered period and spatial area are both huge. As a result, its complexity increases rapidly as well, making the measurement of spatial-temporal dependencies less effective. Therefore, dividing the spatial and temporal information can reduce complexity and obtain better effectiveness and computational efficiency. Second, existing techniques are not designed to measure the entire spatial-temporal dependencies simultaneously. Since most preeminent deep learning techniques focus on processing either spatial information (CNN/GCN) or temporal sequence (LSTM/GRU), it is logical to appoint several of these \say{generals} to divide and conquer the spatial-temporal information. For example, in \cite{stresnet} and \cite{ZHANG_TKDE}, spatial information sampled from different periods (hourly, daily, and weekly) is first measured by multiple deep residual CNNs, then the results are merged by fully connected networks to combine the impacts from different periods. In \cite{stdn} and \cite{st_mgcn}, the spatial information from each timestamps is first measured by CNNs or GCNs. Then the convolutional results enter LSTM as latent spatial representations to calculate the final outputs. Generally, divide-and-conquer is a reasonable strategy to solve problems with high complexity. However, in spatial-temporal prediction, it also divides and distorts the dependencies, which limits the prediction performance.

Besides, when passing through multiple architectures, the spatial and temporal information is condensed and projected into latent representation space. Indeed, the complexity can be therefore reduced, but the spatial-temporal features also turn implicit, making the interpretation of the predicted result difficult. Consequently, current methods are hard to be practically deployed because understanding where the crowd flows come from and being aware of the relationships between predicted result and historical observations are critical for high-level applications.

To overcome these challenges, we propose the \textbf{S}patial-\textbf{T}emporal \textbf{S}elf-\textbf{A}ttention \textbf{N}etwork (\textbf{STSAN}). Instead of divide-and-conquer, we develop an ST encoding gate to calculate the entire spatial-temporal representation with corresponding positional and time encodings. Moreover, in order to measure the entire spatial-temporal dependencies simultaneously and meanwhile maintain a decent efficiency, we propose a Multi-aspect attention mechanism to perform scaled dot-product attention over the spatial-temporal information. Furthermore, the attention weights explicitly calculated for each spatial-temporal position can be extracted for prediction interpretation, which allows the urban manager to troubleshoot correspondingly during practical use.

The contributions of our work can be summarized as follows:
    
\begin{itemize}
    \item In STSAN, we propose the ST encoding gate that represents the entire spatial-temporal observation with the corresponding positional and time information in a complete feature space, which preserves the sophisticated spatial-temporal dependencies for more effective prediction.
    \item We propose a Multi-aspect attention mechanism, which can apply scaled dot-product attention to the entire spatial-temporal information. Moreover, by explicitly attending to every spatial-temporal position, the attention weights can indicate how STSAN understands the historical observation and help to interpret the prediction making.
    \item We extensively evaluate our model on three datasets and demonstrate that it achieves significant error reduction over the state-of-the-art baselines.
\end{itemize}

\section{Related Work}
Recently, deep learning methods have achieved significant improvement in spatial-temporal prediction for urban computing. Since LSTM demonstrated extraordinary effectiveness in processing time-series information, it is adopted to improve the performance of traffic prediction \cite{DBLP:cui_ke_wang}. In the meantime, as researchers noticed that not only time-series information but also spatial dependencies are crucial, the features of surrounding areas are also considered in traffic flow prediction \cite{Zhang:2016:DPM:2996913.2997016}. Thereupon, a plentiful amount of works, including predicting crowd flow \cite{stresnet} and ride-hailing demand \cite{ke_zheng_yang}, started to implement convolutional neural networks to measure spatial dependencies. In order to capture both the spatial and temporal features, different structures, such as merging the convolutional results of multiple periods \cite{stresnet} or feeding each convolutional result into LSTM \cite{ConvLSTM,shi_2017,dmvstnet}, are extensively investigated. In recent works, more sophisticated gating \cite{stdn}, merging \cite{ZHANG_TKDE}, and graph convolution \cite{st_mgcn} mechanisms are proposed to enhance the measurements of spatial and temporal dependencies.

Another trend of spatial-temporal prediction research is based on graph structured data (e.g., highway sensor data) and has been increasingly investigated as well. Enormous works rely on graph convolution \cite{gcn,deep_gcn,chebnet,diff_cnn,gen_gcn} to measure the spatial features of graphs. For instance, DCGRU \cite{dcgru} and LC-RNN \cite{lcrnn} are developed to capture the local spatial dependencies on traffic networks. ST-GCN applies multiple nested convolutional structures in traffic forecasting to extract spatial and temporal features \cite{stgcn}. GSTNet further investigated capturing global dynamic dependencies to improve in traffic network prediction tasks \cite{GSTNet}. LRGCN introduced R-GCN with Long Short-Term Memory and a novel path embedding method for path failure prediction \cite{lrgcn}.

However, in the geographical and graph-based methods mentioned above, divide-and-conquer is still the predominant strategy given the complexity of spatial-temporal information and the limitations of the adopted deep learning techniques. Moreover, since they measure the spatial-temporal information implicitly, the dependencies lead to the predicted results are uninterpretable.

\begin{figure}[t]
    \centering
    \includegraphics[width=0.8\linewidth]{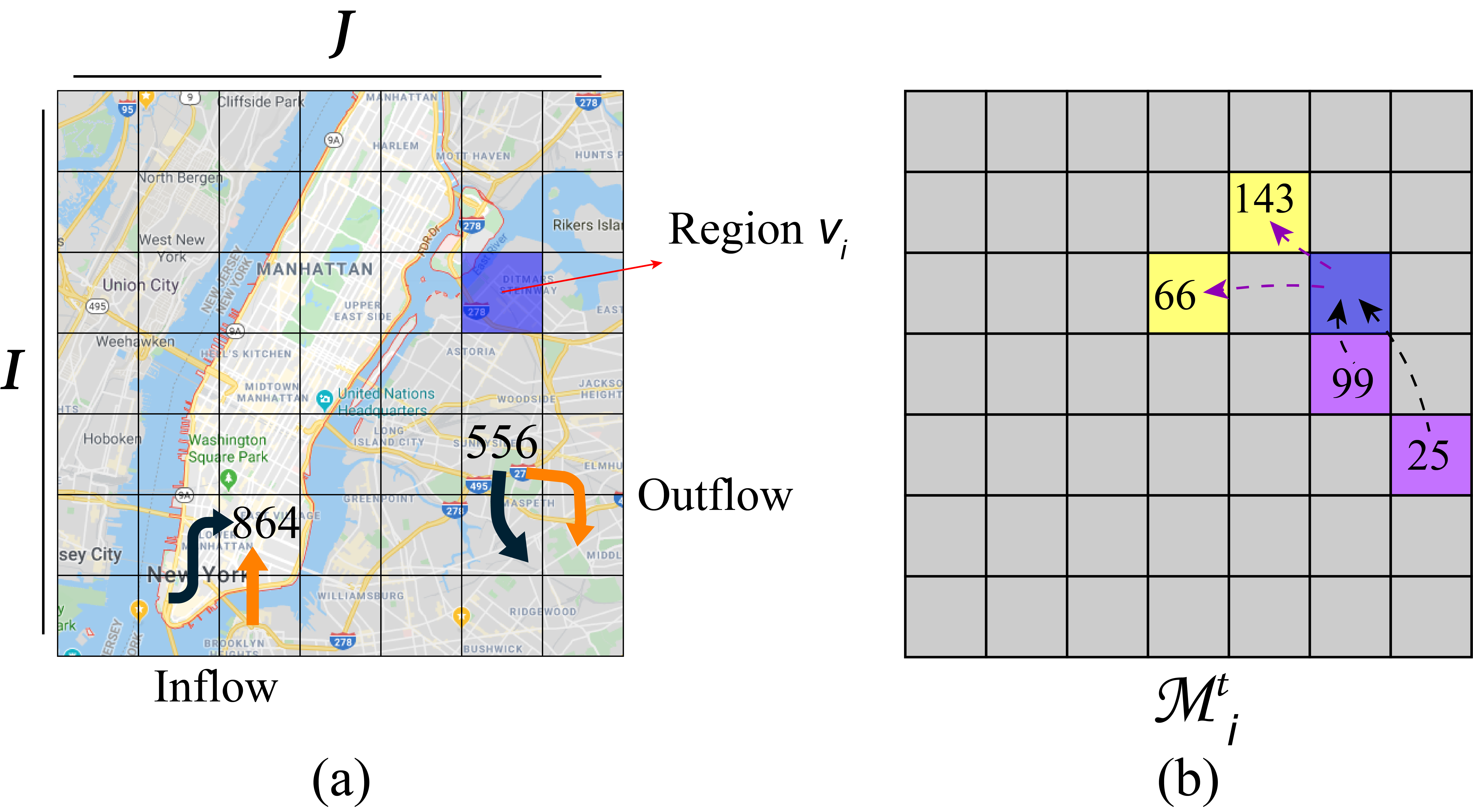}
    \caption{(a) Map segmentation regions' crowd flows. (b) Visualization of transition matrix \(\mathcal{M}_i^t\).}
    \label{fig_prob_def}
\end{figure}
    
\begin{figure*}[t]
    \centering
    \includegraphics[width=0.75\textwidth]{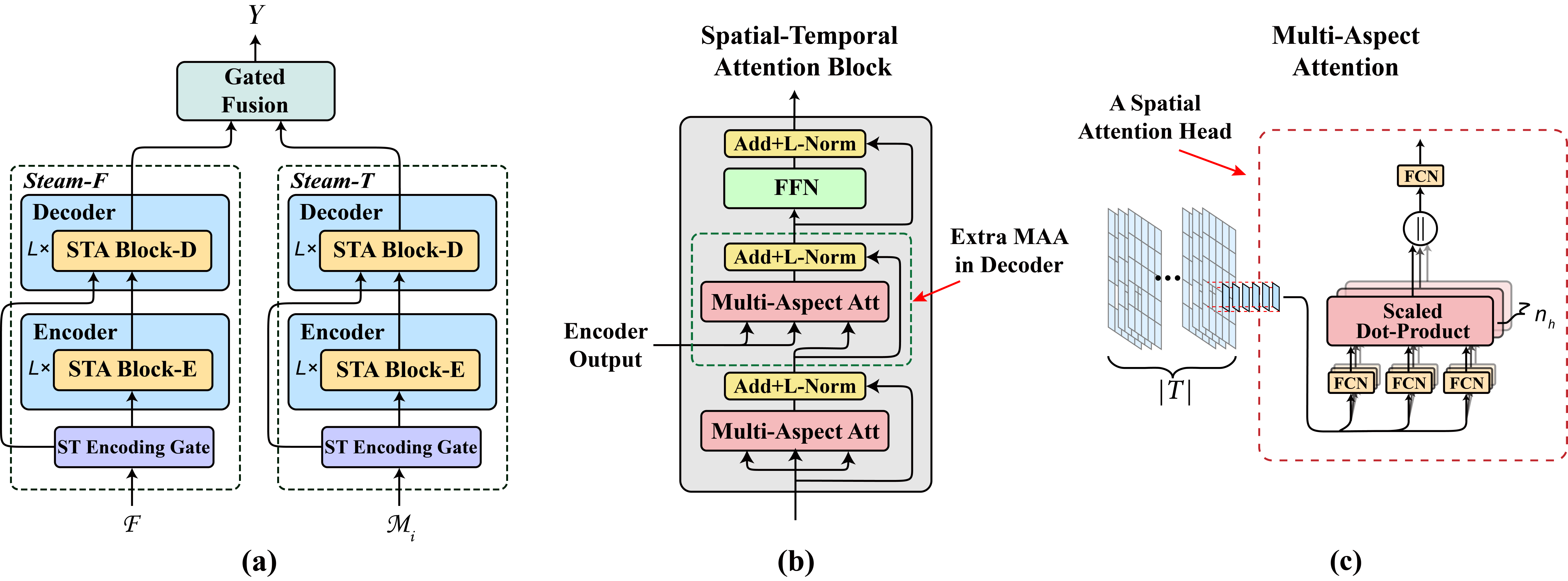}
    \caption{(a) Model architecture. (b) Spatial-Temporal Attention Block. (c) Simplified visualization of a spatial attention head.}
    \label{fig_model_arch}
\end{figure*}
    
\section{Notations and Problem Formulation}
As shown in Figure~\ref{fig_prob_def}, we divide an area into a $\textit{I} \times \textit{J}$ grid map with \textit{N} grids (\textit{N = $I \times J$}). Each grid represents a region, denoted as \{$v_1$, $v_2$, ..., $v_n$\}. $T = \{t_1,t_2,...,t_m\}$ contains all available time intervals of the historical observations. There are $w = 2$ types of features (inflow, outflow) included in each region at one interval. Specifically, when an object (e.g., person, vehicle) was in $v_s$ at time $t_s$ and appeared in $v_e$ at time $t_{s+1}$, it contributed one outflow to $v_s$ and one inflow to $v_e$. The overall inflow and outflow of $v_i$ at interval $t$ are denoted as \( \mathcal{F}_{i,t}^{in} \) and \( \mathcal{F}_{i,t}^{out} \). At the meantime, the transitions between regions are extracted. When $v_i$ is considered, the inflows and outflows between $v_i$ and every other region are calculated explicitly. For example, at interval $t$, \(\mathcal{M}_{i}^{t,in} \in \mathbf{R}^{I \times J}\)  and \(\mathcal{M}_{i}^{t,out} \in \mathbf{R}^{I \times J}\) stand for the in and out transition matrices of $v_i$, where \(\mathcal{M}_{i,j}^{t,in}\) and \(\mathcal{M}_{i,j}^{t,out}\)  indicate the flow volumes from $v_j$ to $v_i$ and from $v_i$ to $v_j$.

\noindent \textbf{Problem Statement} Given historical observations \(\mathcal{F} \in \mathbf{R}^{I \times J \times T \times w}\) and \(\mathcal{M}_i \in \mathbf{R}^{I \times J \times T \times w}\), the crowd flow prediction problem is formulated as learning a function $\textit{f}_\theta$ that maps the inputs to the predicted crowd flows $\hat{Y}_i$ at the upcoming timestamp:
    
\begin{equation}
    \hat{Y}_i = \textit{f}_\theta(\mathcal{F}, \mathcal{M}_i)
\end{equation}
    
\noindent where \(\hat{Y}_i \in \mathbf{R}^w\) and $\theta$ stands for the learnable parameters.

\section{Spatial-Temporal Self-Attention Network}
Figure~\ref{fig_model_arch} (a) illustrates the architecture of STSAN, which consists of two streams of encoder and decoder to measure the flow and transition information independently. At the entry of each stream, an ST encoding gate is developed to calculate the representation. Both the encoders and the decoders contain $L$ spatial-temporal attention block (STA block), where the Multi-aspect attention (MAA) is implemented for self-attention (Figure~\ref{fig_model_arch} (b)). After the two streams, a gated fusion mechanism is implemented to merges their results and generates the predicted crowd flows. The details of each component are discussed in the following subsections.

\subsection{Spatial-temporal Endocing Gate}
Representing the entire spatial-temporal information in a single feature space requires corresponding positional and time encodings to be added to each spatial-temporal position. Such information used to be intrinsically encoded through CNN/GCN and LSTM. However, when the spatial-temporal features are measured merely by attention mechanism, it no longer exists. Therefore, we propose the Spatial-temporal encoding gate (STEG) to calculate the representation with positional and time encodings, which allow the attention mechanism to distinguish the features at different spatial-temporal positions.

\begin{figure}[t]
    \centering
    \includegraphics[width=0.7\linewidth]{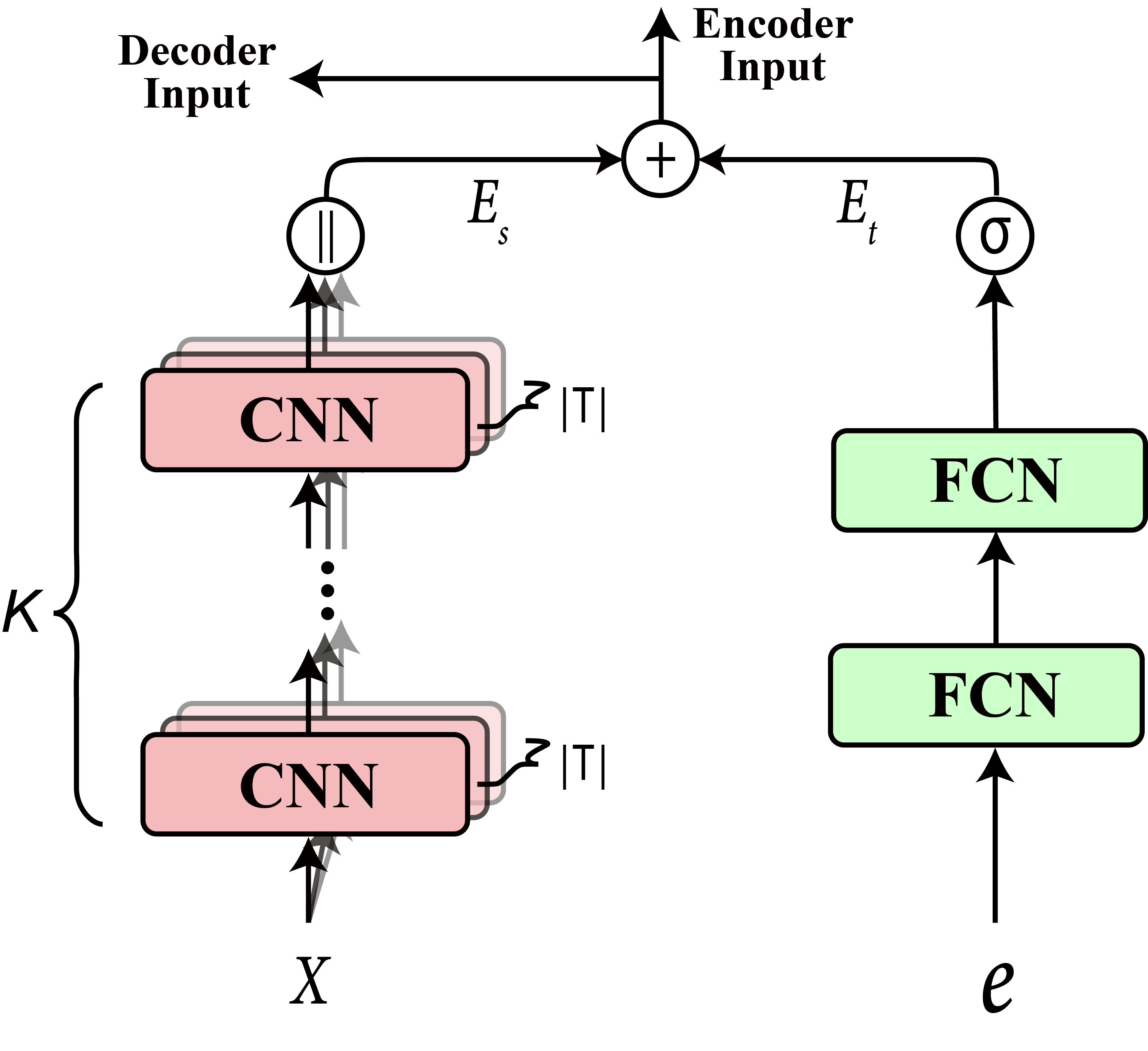}
    \caption{Spatial-Temporal Encoding Gate}
    \label{fig_stpe}
\end{figure}

As shown in Figure~\ref{fig_stpe}, the STEG consists of two parts, which include a spatial encoding network (SEN) on the left and a temporal encoding network (TEN) on the right. The SEN generates the positional encodings for different regions, and the TEN transforms the timestamps with external information into temporal encodings.

In the SEN, $|T|$ stacks of $K$ CNN layers are applied to the historical observation of each timestamp. We denote the input as \(X \in \mathbf{R}^{I \times J \times T \times w}\) (e.g., \(\mathcal{F}\) or \(\mathcal{M}_i\)), and $X_t \in \mathbf{R}^{I \times J \times w}$ is a temporal slice of $X$. Each layer of a CNN stack applies convolution operation to $X_t$:

\begin{equation}
    f(X_t) = ReLU(X_t \ast_c W_c)
\end{equation}

\noindent where \(W_c \in \mathbf{R}^{k \times k \times d}\) is a convolution kernel that contains $d$ filters with $k = 3$ as the filter length, and $\ast_c$ denotes the convolution operation. Notice that $d$ is also the common feature dimension shared across STSAN. The CNNs correlate a region with its neighbors and encode it into a unique representation in a higher-dimension feature vector. In order to maintain the matrix shape, zero padding is adopted during the convolutions. After each CNN stack finishes the $K$ convolutions, \(E_s \in \mathbf{R}^{I \times J \times T \times d}\) will be formed as:

\begin{equation}
    E_s = ||_{t=1}^{|T|} f(X_t)^K
\end{equation}

\noindent where $||$ indicates the concatenation operation.

To generate the temporal encoding, we first represent the timestamps, which consist of day-of-week and time-of-day information, in a $\mathbf{R}^{|T| \times (7+P)}$ one-hot matrix, where P is the number of time intervals in one day. Besides, $z$ types of external information (e.g., temperature, rainfall, holiday), are recorded in a $\mathbf{R}^{|T| \times z}$ matrix. Then the two matrices are concatenated to form $e \in \mathbf{R}^{|T| \times (7 + P + z)}$, which is later transformed into $E_t \in \mathbf{R}^{|T| \times d}$ by a two-layer fully-connected network (FCN).

After $E_s$ and $E_t$ are generated, the spatial-temporal representation $H \in \mathbf{R}^{I \times J \times |T| \times d}$ encoded with positional and time information are produced by $H = E_s + E_t$, where $E_t$ is broadcasted to the shape of $E_s$. Notice that the encoded spatial-temporal representation at the latest timestamp is extracted as the decoder input.

\subsection{Multi-Aspect Attention}
Based on Multi-head attention (MHA) \cite{transformer}, we further propose Multi-aspect attention (MAA) to perform the scaled dot-product attention function over spatial-temporal information. An attention function can be described as mapping a query and a set of key-value pairs to an output, where typically the query, keys, values, and output are words represented as vectors. In MHA, the original word features are split and measured with multiple attention heads, which allows the model to jointly attend to information from different representation subspaces at different positions. When performing attention function in a spatial-temporal feature space, the inputs have an extra spatial realm. Therefore, in MAA (Figure~\ref{fig_model_arch} (c)), we propose to treat each of the spatial dimension (e.g., length, width) as a subspace and further allocate spatial attention heads to measure each position explicitly.

We denote $Q \in \mathbf{R}^{I \times J \times |T_Q| \times d}$, $K \in \mathbf{R}^{I \times J \times |T_{K}| \times d}$, and $V \in \mathbf{R}^{I \times J \times |T_{K}| \times d}$ as the inputs of the MAA, in which the multi-head scaled dot-product is first performed by each spatial head over $Q_{i,j}^h \in \mathbf{R}^{|T_Q| \times d_h}$, $K_{i,j}^h \in \mathbf{R}^{|T_{K}| \times d_h}$, $V_{i,j} \in \mathbf{R}^{|T_{K}| \times d_h}$, where $h$ indicates the $h$-th of $n_h$ feature heads and $d_h$ = $\frac{d}{n_h}$:

\begin{equation}
    s_{i,j}^h =  \frac{Q_{i,j}^h \cdot (K^h_{i,j})^T}{\sqrt{d_h}}
\end{equation}

\noindent Next, a softmax function is applied on $s^h \in \mathbf{R}^{I \times J \times |T_Q| \times |T_K|}$ to calculate the weighted attentions:

\begin{equation}
    \alpha_{i,j,t}^h = \frac{\exp(s^h_{i,j,t})}{\sum_{i=1}^I \sum_{j=1}^J \sum_{t=1}^{|T|} \exp(s^h_{i, j, t})}
\end{equation}

\noindent Then, dot-products between the attention weights and the value $V_{i,j}$ are performed by each feature head, whose results are later concatenated to generate the overall output of $head_{i,j}$:

\begin{equation}
    h(Q_{i,j},K_{i,j},V_{i,j}) = ||_{h=1}^{n_h}\alpha_{i,j}^h \cdot V_{i,j}^h
\end{equation}

\noindent Since all spatial heads can be computed in parallel, the learning and inference are computationally efficient, although the entire feature space is large. Finally, the MAA can be expressed as a function aggregating the results of all spatial heads:

\begin{equation}
    \begin{aligned}
        MA&A(Q, K, V) =\Big(||_{i=1}^I||_{j=1}^J h(Q'_{i,j}, K'_{i,j}, V'_{i,j})\Big)W^O \\
        &where\ Q' = QW^Q,\ K' = KW^K,\ V' = VW^V
    \end{aligned}
\end{equation}

\noindent
where $W^Q \in \mathbf{R}^{d \times d}, W^K \in \mathbf{R}^{d \times d}, W^V \in \mathbf{R}^{d \times d}, W^O \in \mathbf{R}^{d \times d}$ are learned linear transformation matrices.

In STSAN, the MAA is used in two different ways: (1) The first layer of the STA block is a self-attention MAA, where the queries, keys, and values are generated from the same input, which is the output of the previous layer. Self-attention allows each position in the spatial-temporal representation to attend to all positions in the output of the previous layer. (2) There is an additional MAA layer in the STA block of the decoder, where the queries come from the previous layer, and the keys and values come from the output of the encoder. In this way, every position in the decoder can attend over all positions in the historical observations from the encoder. This mimics the encoder-decoder attention mechanisms in sequence-to-sequence language models \cite{NMTJL,seq2seq}.

\begin{table*}[t]
    \centering
    \begin{tabular}{ l||c|c||c|c||c|c }
        \hline \hline
        \multirow{2}{*}{Model} & \multicolumn{2}{c||}{Taxi-NYC} & \multicolumn{2}{c||}{Bike-NYC} & \multicolumn{2}{c}{Mobile M} \\
        \cline{2-7}
        & inflow & outflow &  inflow & outflow & inflow & outflow \\
        \cline{2-7}
        \hline \hline
        HA & 45.19/24.94 & 53.71/32.09 & 20.15/13.04 & 20.39/13.04 & 52.32/28.11 & 52.13/27.97 \\
        ARIMA & 33.54/18.62 & 40.70/23.61 & 17.14/10.83 & 18.03/11.28 & 35.62/20.74 & 34.95/20.42 \\
        VAR & 48.04/23.21 & 128.67/29.84 & 27.37/14.29 & 27.67/15.09 & 48.12/24.38 & 67.01/34.20 \\
        MLP & 27.33/16.99 & 32.88/20.78 & 10.94/7.64 & 11.73/7.86 & 29.48/20.14 & 29.41/20.02 \\
        LSTM & 24.40/15.10 & 30.47/19.20 & 11.55/8.07 & 12.59/8.52 & 28.79/19.82 & 28.01/19.46 \\
        GRU & 24.35/15.17 & 30.33/19.19 & 11.71/ 8.21 & 12.49/8.32 & 28.54/19.59 & 28.14/19.68 \\
        ST-ResNet & 20.22/12.83 & 25.47/16.18 & 9.24/6.70 & 10.42/7.31 & 25.81/16.95 & 25.94/17.04 \\
        DMVST-Net & 18.91/12.18 & 24.01/15.34 & 8.99/6.54 & 9.75/6.84 & 24.73/15.86 & 24.78/15.93 \\
        STDN & 17.93/11.38 & 23.43/14.87 & 8.57/6.25 & 9.47/6.63 & 24.37/15.51 & 24.45/15.66 \\
        \hline
        Transformer & 22.38/13.98 & 27.17/17.84 & 10.39/7.71 & 11.73/7.89 & 27.62/19.04 & 27.44/18.92 \\
        STSAN w/o STEG & 18.73/11.91 & 23.79/15.27 & 8.92/6.50 & 9.90/6.91 & 25.04/16.16 & 25.15/16.19 \\
        STSAN & \textbf{15.13}/\textbf{9.82} & \textbf{22.75}/\textbf{14.15} & \textbf{7.27}/\textbf{5.22} & \textbf{8.91}/\textbf{6.17} & \textbf{23.93}/\textbf{15.29} & \textbf{23.80}/\textbf{15.38}\\
        \hline \hline
    \end{tabular}
    \caption{Experimental Results (RMSE/MAE). Transformer: the self-attention method without ST encoding gate and Multi-aspect attention. STSAN w/o STEG: STSAN without ST encoding gate. ST-MGCN is not included since its code is not released.}
    \label{tbl_results}
\end{table*}

\subsection{Gated Fusion}
A gated fusion mechanism is proposed to merge the outputs of the Stream-T and the Stream-F and generate the final prediction. It contains $K_f$ fusion layers, which can be defined as:

\begin{align}
    &O_t^l = ReLU(O_t^{l-1} \ast_c W_t^l)\\
    O_f^l &= ReLU(O_f^{l-1} \ast_c W_f^l) \circ \sigma(O_t^l)
\end{align}

\noindent
where $W_t^l$ and $W_f^l$ are the convolution kernels of the $l$-th layer, and $\sigma$ denotes the sigmoid activation function. We denote the outputs of Stream-F and Stream-T as $O_f^0$ and $O_t^0$. In each fusion layer, sigmoid activation is applied to the convolutional result $O_t^l$, which is then transformed into a gated matrix and further multiplied with the convolutional result of $O_f^{l-1}$ through Hadamard product $\circ$. Finally, the output $O_f$ of the last fusion layer is flattened, denoted as $O_{flat} \in \mathbf{R}^{1 \times d_f}$, and fed into a fully connected network:
    
\begin{equation}
    \hat{Y}_i = Tanh(O_{flat}W + b)
\end{equation}
    
\noindent where $W \in \mathbf{R}^{d_f \times w}$ and $b$ are the learnable parameters while $Tanh$ denotes the $tanh$ activation function.

\subsection{Encoder and Decoder}
The STSAN employs an encore-decoder structure \cite{seq2seq}, where the encoder calculates the continuous representation of spatial-temporal historical observations, and the decoder generates the output via performing Multi-aspect attention over the decoder query and the encoder outputs.

As shown in Figure~\ref{fig_model_arch} (a), the encoders are composed of a stack of $L$ STA blocks, where the Multi-aspect attention mechanism (MAA) is followed by a feed-forward network (FFN). The feed-forward network is a two-layer fully connected network: $f(x) = ReLU(xW_1 + b_1)W_2 + b_2$, which can be considered as two convolutions with kernel size 1 to further correlates the spatial-temporal outputs of the MAA. Besides, residual connection \cite{DeepResidual} is adopted after each sub-layer as well as layer normalization \cite{layernorm}. The output of each sub-layer can be defined as \(LayerNorm(x + f(x))\), where $f$ is the implemented function of the sub-layer. To facilitate this construction, the outputs of the sub-layers have an identical feature dimension $d$.

The decoder is also composed of a stack of $L$ STA blocks. In addition, another MAA layer is added in the middle of the STA block to perform the traditional attention mechanism over the self-attention results and the encoder outputs (Figure~\ref{fig_model_arch} (b)). Residual connection and layer normalization are also applied to the output of each sub-layer.

STSAN is trained by minimizing the mean-square-error between predicted results and ground truths:
    
\begin{equation}
    \mathcal{L}(\theta) = \frac{(Y_i - \hat{Y}_i)^2}{w}
\end{equation}

\noindent where $\theta$ denotes the learnable parameters of STSAN.
    
\section{Experiment}
\subsection{Datasets}
We evaluate our model on three datasets -- Taxi-NYC, Bike-NYC, and Mobile M (Table~\ref{tbl_datasets}). Taxi-NYC and Bike-NYC contain 60 days of trip records, which includes the locations and times of the start and the end of a trip. We use the first 40 days as training data and the rest 20 days as testing data.
Since Taxi-NYC and Bike-NYC contain only the start and end points instead of trajectories, we further acquire Mobile M that contains trajectories of mobile users provided by a service provider. The 90-day data is split to 60 and 30 days for training and testing.

For Taxi-NYC and Bike-NYC, the time interval is set as 30 minutes, which is slightly longer than the average trip duration. For Mobile M, the time interval length is 15 minutes, which is equal to the sampling rate of mobile records. The grid size in Taxi-NYC and Bike-NYC is $1km \times 1km$ while in Mobile M it is $200m \times 200m$ since Mobile M has a larger amount of data within a smaller area. We randomly select 20\% of the training samples for validation and the rest for training.
    
\subsection{Evaluation Metrics \& Baselines} We compare STSAN and its variants with nine baselines based on two metrics: (1) Rooted Mean Square Error (RMSE) and (2) Mean Absolute Error (MAE).
    
\begin{table}[t]
    \centering
    \begin{tabular}{ c|c|c|c }
        \hline \hline
        Datasets & Taxi-NYC & Bike-NYC & Mobile M \\
        \hline
        Grid map size & $16 \times 12$ & $14 \times 8$ & $8 \times 11$ \\
        \hline
        Time interval & 30 mins & 30 mins & 15 mins \\
        \hline
        \multirow{2}{*}{Time Span} & 1/1/2016 - & 8/1/2016 - & 10/1/2018 - \\
        & 2/29/2016 & 9/29/2016 & 12/29/2018\\
        \hline
        Total records & 22,437,649 & 9,194,087 & 158,742,004 \\
        \hline \hline
    \end{tabular}
    \caption{Details of the datasets}\smallskip
    \label{tbl_datasets}
\end{table}

\subsubsection{Baselines}
\textbf{(1)HA}: Historical average; \textbf{(2)ARIMA}: Auto-regressive integrated moving average model; \textbf{(3)VAR}: Vector auto-regressive model; \textbf{(4)MLP}: Multi-layer perceptron. Hidden units: \{16, 32, 64, 128\}, learning rate: \{0.1, 0.01, 0.001, 0.0001\}. The best setting: \{64, 0.001\}; \textbf{(5)LSTM}: Long-Short-Term-Memory. We evaluate multiple hyperparameters: previous frame length in \{3, 6, 12\}, hidden units in \{32, 64, 128\}, and learning rate in \{0.1, 0.01, 0.001, 0.0001\}. We observed that the best setting is \{6, 64, 0.001\}; \textbf{(6)GRU}: Gated-Recurrent-Unit network \cite{DBLP:journals/corr/ChungGCB14}. The hyperparameters are the same from LSTM; \textbf{(7)ST-ResNet}: Spatial-Temporal Residual Convolutional Network \cite{stresnet}; \textbf{(8)DMVST-Net}: Deep Multi-View Spatial-Temporal Network \cite{dmvstnet}. \textbf{(9)STDN}: Spatial-Temporal Dynamic Network \cite{stdn}.

\noindent
We use Adam \cite{adam} as the optimizer for all baselines. For ST-ResNet, DMVST-Net, and STDN, the hyperparameters remain as the optimized settings introduced by their authors.

\subsection{Data Preprocessing}
We use Min-Max normalization to convert both flow and transition volumes to scale of [0, 1] during the training. When sampling the historical observations, we follow the periodic shifting rule introduced in \cite{stdn} and select $P=3$ intervals around the same timestamp of the previous $D=7$ days together with the current interval before the predicted timestamp. We also evaluate tailoring the inputs by limiting the spatial area within a $B \times B$ local block around the predicted region $v_i$. Since regions far away from $v_i$ are usually irrelevant, they may introduce noises into the prediction. If $v_i$ is near the margin, zero padding is adopted to fill the vacant positions. Compared with feeding the global inputs, the tailoring strategy empirically achieves better performance in STSAN and other deep learning baselines. During the evaluation, we filter out all samples whose ground truths are less than ten since values close to zero are easy to predict. As a common criterion, the filtering is applied to all baselines as well.

\subsection{Hyperparameters}
We tune STSAN on the validation set, and observe that $L = 4$, $d = 64$, $n_h = 8$, $K = 3$, $K_f = 2$, $B = 7$, and dropout rate $r_d = 0.1$ achieve the best performance. Adam optimizer is used with warm-up learning rate as introduced in \cite{transformer}. We also tested warm-up on other baselines and observed no improvement. Using the hyperparameters described above, it takes around 5 hours to train our model on one machine with 8 NVIDIA RTX2080Ti GPUs and 1024 as batch size.

\subsection{Results}
As shown in Table~\ref{tbl_results}, traditional statistic methods (HA, ARIMA, and VAR) are significantly less effective. It exposes the weakness of methods that exclusively capture the patterns of historical statistic values and ignore the complicated spatial-temporal dependencies. Among traditional neural networks, MLP merely learns the linear transformation from historical observation to predicted values, where the non-linear spatial-temporal dependencies are omitted. LSTM and GRU obtain considerable improvement compared to traditional time-series methods, given their effectiveness on modeling temporal dependencies. Nonetheless, as the spatial information is not included, their performance is limited.
    
Deep learning methods show significant advantages in capturing complicated spatial-temporal dependencies. ST-ResNet employs three stacks of deep residual network to capture spatial dependencies from three different periods. However, the convolutional results are indiscriminately merged by fully connected networks, which overlooks the distinctive impacts of temporal dependencies. DMVST-Net and STDN show the remarkable capability of modeling both spatial and temporal dependencies through integrating CNNs and LSTMs. However, dividing the measurements of spatial and temporal information also damages the sophisticated spatial-temporal dependencies, which limits their performance. Given its capability of representing and attending to the entire spatial-temporal information, STSAN shows significant improvement compared to the previous deep learning methods.

We also evaluate the effectiveness of the ST encoding gate and Multi-aspect attention. Transformer obtains poor performance as the spatial-temporal information is treated as a sequence. Although \textbf{STSAN w/o STEG} utilizes the Multi-aspect attention mechanism, the positional and time information of the spatial-temporal features is missing, for which the MAA can not distinguish the impacts of different regions and times.

\begin{figure}[t]
    \centering
    \includegraphics[width=0.8\linewidth]{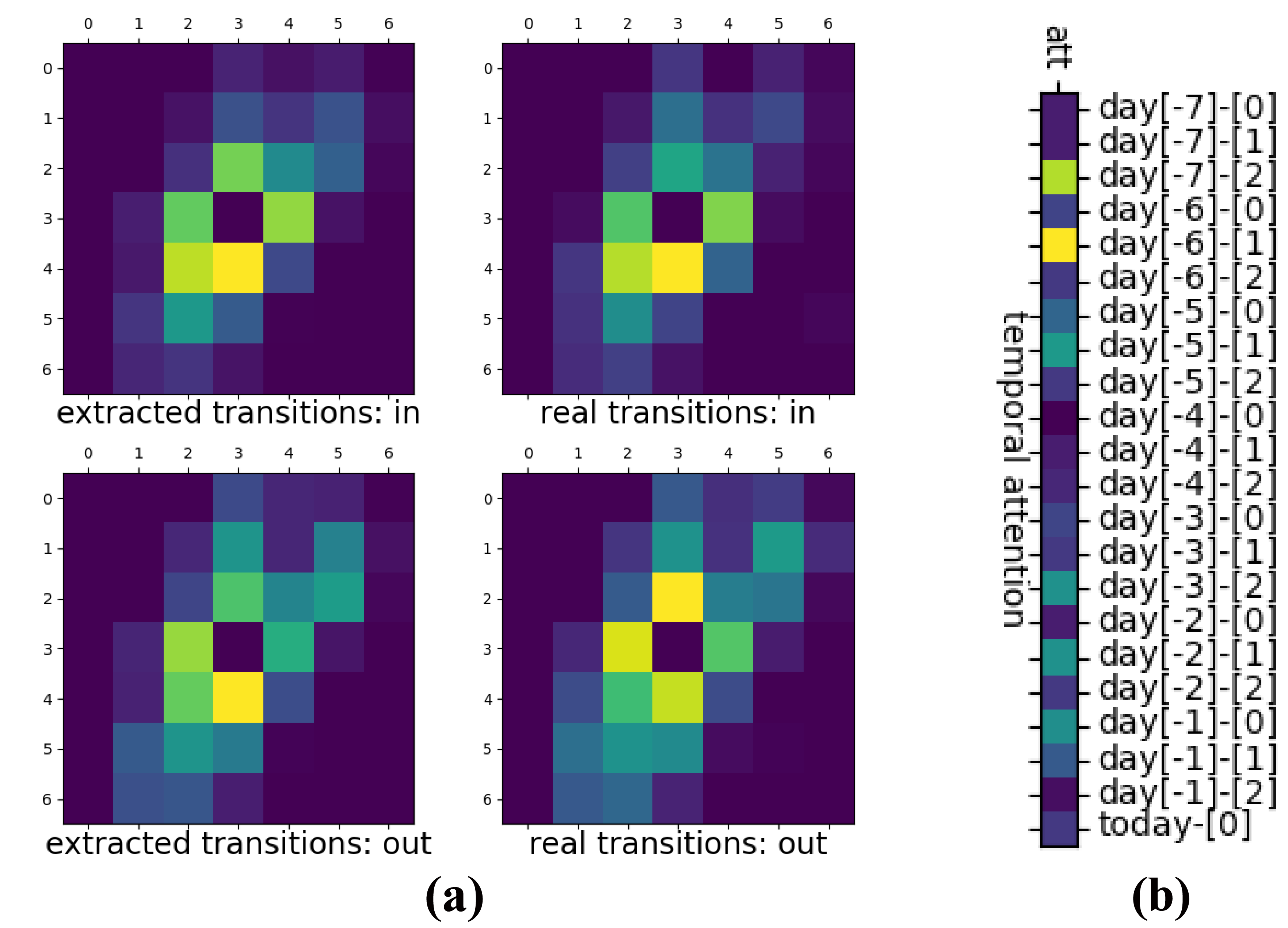}
    \caption{(a) The extracted transitions (left) and the ground truths (right). (b) The temporal attention weights of every historical timestamp. A block is brighter if its value is larger.}
    \label{fig_att}
\end{figure}

\subsection{Interpretation of Crowd Flow Prediction}
We extract the transition features at the end of Stream-T and the attention weights from Stream-F to interpret the predicted results. As shown in Figure~\ref{fig_att} (a), we take out a slice of the transition matrix at the latest timestamp, which is very close to the real transitions in the future. Besides, the temporal attention weights also illustrate which historical moment is most relevant to the predicted timestamp.

\section{Conclusion and Future Work}
In this work, we present the Spatial-Temporal Self-Attention Network for crowd flow prediction. Specifically, an ST encoding gate is developed to represent the entire spatial-temporal information with positional and time encodings. Moreover, we propose a Multi-aspect attention mechanism that applies scaled dot-product attention over the spatial-temporal representation via allocating extra spatial attention heads on every position. Furthermore, the transition features and the attention weights of STSAN can be extracted for prediction interpretation. In the future, we will focus on covering the entire time-evolving graph information and achieving accurate long-term prediction.

\bibliography{ijcai20.bib}
\bibliographystyle{named}

\end{document}